# Technology as uncharted territory: Contextual integrity and the notion of AI as new ethical ground

Alexander Martin Mussgnug / alexander@mussgnug.de
Working paper as of January 2025
**Please cite version published in Philosophy & Technology**

**Abstract:** Recent research illustrates how AI can be developed and deployed in a manner detached from the concrete social context of application. By abstracting from the contexts of AI application, practitioners also disengage from the distinct normative structures that govern them. Building upon Helen Nissenbaum's framework of contextual integrity, I illustrate how disregard for contextual norms can threaten the integrity of a context with often decisive ethical implications. I argue that efforts to promote responsible and ethical AI can inadvertently contribute to and seemingly legitimize this disregard for established contextual norms. Echoing a persistent undercurrent in technology ethics of understanding emerging technologies as *uncharted moral territory*, certain approaches to AI ethics can promote a notion of AI as a novel and distinct realm for ethical deliberation, norm setting, and virtue cultivation. This narrative of AI as new ethical ground, however, can come at the expense of practitioners, policymakers and ethicists engaging with already established norms and virtues that were gradually cultivated to promote successful and responsible practice within concrete social contexts. In response, I question the current narrow prioritization in AI ethics of moral innovation over moral preservation. Engaging also with emerging foundation models, I advocate for a moderately conservative approach to the ethics of AI that prioritizes the responsible and considered integration of AI within established social contexts and their respective normative structures.

## I Introduction

AI is employed in a wide range of contexts — scientists leverage AI in their research, media outlets use AI in journalism, and doctors adopt AI in their diagnostic practice. Yet many have noted how AI systems are often developed and deployed in a manner that prioritizes abstract technical considerations which are disconnected from the concrete context of application. This also results in limited engagement with established norms that govern these contexts. When AI applications disregard entrenched norms, they can threaten the integrity of social contexts with often disastrous consequences. For example, medical AI applications can defy domain-specific privacy expectations by selling sensitive patient data, AI predictions can corrupt scientific reliability by undermining disciplinary evidential norms, and AI-generated journalism can erode already limited public trust in news outlets by skipping journalistic best practices.

This paper argues that efforts to promote responsible and ethical AI can inadvertently contribute to and seemingly legitimize this disregard for established contextual norms. Echoing a persistent undercurrent in technology ethics of understanding emerging technologies as *uncharted moral territory*, certain approaches to AI ethics can promote a notion of AI as a novel and distinct realm for ethical deliberation, norm setting, and virtue cultivation. This narrative of AI as new ethical ground, however, can come at the expense of practitioners, policymakers, and ethicists engaging with already established norms and virtues that were gradually cultivated to promote successful and responsible practice within concrete social contexts. In response, this paper questions the current prioritization in AI ethics of moral innovation over moral conservation.

My argument proceeds in four parts. Building upon Helen Nissenbaum's framework of contextual integrity, section two illustrates how AI practitioners' disregard for cultivated contextual norms can threaten the very integrity of contexts such as mental health care or international development. Section three outlines how a tendency to understand novel technologies as uncharted ethical territory exacerbates this dynamic by playing into and seemingly legitimizing disregard for contextual norms. In response, section four highlights recent scholarship that more substantially engages with the contextual dimensions of AI. Under the label of *"integrative AI ethics,"* I advocate for a moderately conservative approach to such efforts that prioritizes the responsible and considered integration of AI within established social contexts and their respective normative structures. After addressing some possible objections and engaging with emerging work on foundation models, I conclude in section five.

## II AI Practice and Contextual Integrity

Several scholars have noted how AI development and deployment can take place in a manner detached from the concrete domain of application. Birhane et al. (2022), for instance, have surveyed influential machine learning conference publications and shown how abstract technical considerations often dominate over the social and ethical dimensions of applications. Mussgnug (2022) has outlined how approaching supervised machine learning generically as a class of "prediction problems" can circumvent engagement with context-specific considerations. The abstractive practices of AI development are also the motivating concern for sociotechnical scholarship on AI. Sociotechnical research problematizes how AI systems are often narrowly conceived in technical terms without acknowledgement of how AI operates through the complex entanglement of social and technological aspects (Kudina & van de Poel, 2024; Selbst et al., 2019).

This section focuses on how abstracting from the concrete domains of application also leads practitioners to disengage from the contextual norms that govern those domains. This disregard for contextual norms can threaten the very integrity of a context with decisive ethical implications. In exploring this dynamic, I draw from Helen Nissenbaum's influential framework of contextual integrity, originally proposed to make sense of privacy debates following the popularization of internet services.[1]

Nissenbaum contends that existing accounts of privacy (in particular the distinction between public and private information) fail to track people's expectations and concerns. In response, she proposes the notion of contextual integrity, not as a definition of privacy but as a framework to help us understand the source of public objection to emerging informational practices and as an initial yardstick for their potential legitimacy. Her approach rests upon the observation that "finely calibrated systems of social norms, or rules, govern the flow of personal information in distinct social contexts" (Nissenbaum, 2009, p. 3). These contextual informational norms have been cultivated over time and in ways sensitive to the particularities of the context in order to "define and sustain essential activities and key relationships and interests, protect people and groups against harm, and balance the distribution of power" (ibid, p.3). Where information technologies defy these established informational norms, Nissenbaum speaks of a prima facie violation of contextual integrity.

[1]The framework of conceptual integrity emerges from a sociotechnical perspective and its emphasis on understanding technological systems not in isolation but as partly constituted by the social context they are employed in (Nissenbaum, 2010). As I will illustrate later, it can also provide a critical perspective on particularly those (misled) approaches to sociotechnical design and ethics that overemphasize the novelty of AI as a sociotechnical system and the need for designing new norms and virtues absent substantial engagement with established contextual norms and virtues.

A prima facie violation of contextual integrity is not sufficient for something to be deemed morally objectionable. Entrenched norms can be misguided or harmful and the implementation of novel technologies in an existing context might call us to question them. Sometimes we might employ new technologies precisely with the intention to overhaul existing moral customs. Nissenbaum concedes that we might challenge entrenched norms in favor of nonconforming practices, "when the latter are shown more effective in supporting or promoting respective [context-based] values, ends, and purposes." (2009, p. 181). At the same time, the framework of contextual integrity relies on a measured conservatism grounded in an acknowledgement that established informational norms often reflect the cultivated wisdom, accumulated experience, and settled rationale of a context. Violations of these norms, thus, should initially alarm us and call for us to critically scrutinize their implications.

We can tie Helen Nissenbaum's framework of contextual integrity to existing criticisms of AI abstraction. As AI practitioners abstract from the concrete context of application, they also detach from the informational norms that govern these contexts. Failing to engage with and, as a result, flouting these informational norms threatens the integrity of a context. In this way abstractive practices in AI contribute to the many egregious privacy breaches witnessed in recent years. However, it is not only informational norms that are regularly disregarded by AI practitioners. In the same way that abstractive practices can contribute to the flouting of informational norms, they can lead practitioners to overlook a broader set of context-specific norms in the form of best practices, established procedures, methodological and epistemic requirements, standards of excellence, and cultivated virtues.

This broader set of contextual norms, much like informational norms, are often the result of cultivated wisdom, accumulated experience, and lessons learned from past failures. Together, they coordinate actors and their expectations, provide protection against harm, and generally aim to promote the functioning of a context in line with certain epistemic and ethical aims. Thus I argue that the integrity of a context rests not only on its informational norms but also on a broader set of contextual norms including best practices, methodological and epistemic requirements, established procedures, and standards of excellence.[2]

The remainder of this section exemplifies this point by examining AI applications in mental health care and international development. I focus on two main issues in each domain, showing how AI applications fail to meet informational norms and diagnostic best practices in mental health care and violate methodological standards and interpretability requirements in international development. The case studies highlight how disregard for contextual norms can threaten the integrity of mental health care and international development — eroding trust, inflicting harm, and sparking justified social outcry.

### II.1 Mental Health Care

Mental health care is governed by a particularly rich set of context-specific norms. Many of those norms are inherited from clinical care and biomedical research more generally. These include those codified in law, such as informational norms pertaining to the treatment and storage of "protected health information" (HIPAA, Pub. L. No. 104-191) or ethical principles laid out in official documents, such as the Nuremberg Code (US v.

[2] Social contexts can be looked at in a wide range of ways – for instance, based on constitutive rules (e.g., Searle, 1996), social goods (Walzer, 1983), or practices (e.g., Schatzki, 2000). I follow Helen Nissenbaum in not endorsing a particular notion of context, mostly because I believe that intuitions and exemplars get us quite far but also since defending a substantive account of contexts is an endeavor not feasible within the scope of this paper. What is more, endorsing one account is likely overly restrictive and misaligned with the aims of this article. The notion of context that proves fruitful is the one which we have established rich and robust norms. I assume not only that this might differ between instances but also that there are cases where a detailed analysis must look at various differently defined contexts at different levels of abstraction to arrive at a complete picture (c.f., Nissenbaum, 2009, Chapters 7 & 9).

Brandt, 1947) or the Belmont Report (1978).[3] These principles were developed in response to egregious abuse in German concentration camps or during the Tuskegee Syphilis Study and provide a standard against which practitioners today can be held responsible. In addition, professional organizations such as the American Psychological Association (2017) or the Royal College of Psychiatrists (2014, 2017) have adopted designated codes of ethics, which attempt to outline the values, virtues, and best practices of mental health care in particular. Psychological and psychiatric practice has, over the past century, also established epistemic and methodological standards. For instance, the Diagnostic and Statistical Manual of Mental Disorders and the International Classification of Diseases attempt to provide taxonomies of psychological and psychiatric conditions and benchmarks for their diagnosis.

These and other contextual norms do not guarantee that mental health care always proceeds smoothly and ethically. Far from it: best practices can be misguided, principles leave much leeway for (mis)interpretation, values and incentives can be misaligned. Nonetheless, these context-specific procedures, principles, virtues, standards, and best practices have often been put in place for good reason. They coordinate actors and their expectations. And their considered adherence in line with the cultivated skill and practical wisdom of practitioners generally leaves mental health care better off than general dismissal or ignorance of such norms.[4] In what follows, I briefly outline how AI applications can breach informational norms and best practices with decisive implications for the integrity of mental health care.

*Mental health chatbots and privacy:*

Numerous companies have developed AI chatbots promising prompt mental health advice, at a time where qualified therapists are costly and in short supply. Meanwhile, sensitive data from applications where users discuss their most personal circumstances and struggles is shared rather widely, for instance for targeted marketing. An extensive report by the Mozilla Foundation (2022) describes these practices as nothing less than a "data harvesting bonanza." Even the nonprofit organization Crisis Text Line has faced scrutiny for sharing user data with its for-profit spin off customer service AI (Levine, 2022).

The notion of contextual integrity provides one way of understanding these privacy breaches. Providers introduce AI chatbots, promising solutions and services comparable to traditional therapy and support. Sharing information widely with commercial actors and without meaningful informed consent, however, breaches the informational norms that govern therapeutic relationships in the mental health context. In the same vein, privacy regulations that intend to protect sensitive health care data such as HIPAA are not enforced when it comes to mental health AI systems (Subbian et al., 2021). As AI applications breach entrenched informational norms and the respective expectations regarding privacy that patients have, they threaten the integrity of mental health care. They erode trust, cause harm and spark justified social outcry.

[3] See also the Declaration of Helsinki (World Medical Association, 2013) or the widely adopted principles of bioethics formulated by Beauchamp and Childress (2013).

[4] The integrity and functioning of mental health care, however, rests not only on such norms but on their skillful consideration and application by those involved in the field. Codified principles might conflict when considering alternative courses of care, standardized categories might leave room for ambiguity when diagnosing a particular patient, and best practices might prove infeasible in certain on-the-ground circumstances. It is here that practitioners in the context of mental health care must exercise judgment and skill. The functioning and integrity of mental health care rests on a virtuous pairing of practical wisdom and established norms. On the one hand, practical wisdom is required for the application and development of suitable norms. On the other hand, norms and regulations can encourage the cultivation of practical wisdom, for instance by requiring mandatory ethical training, guided clinical experience, and other pedagogical practices.

*AI applications and diagnostic best practice*

AI applications in mental health care also flout diagnostic best practices. Key to psychiatric practice is the process of differential diagnosis. Mental health professionals iteratively eliminate possible conditions before settling on a diagnosis to ensure that no condition is missed or misattributed (First, 2024; Frances, 2013). When diagnosing a patient, such methodological best practices must be paired with skillful judgment. Practitioners must weigh between the access to care and insight that an early diagnosis enables and the potential reinforcing effects of labeling (Becker, 1963) or the risk of a misdiagnosis (Frances, 2013, p. 6). Clinicians must consider how a patient's culture shapes their response to a diagnosis or how differences in lived experiences between them and their patients play out in the diagnostic relationship (American Psychiatric Association, 2013, p. 14). Diagnostic practice is, thus, not just an exercise of checking off symptoms. Instead, it demands a great deal of situational awareness and care (Kroll & Mason, 2021).

Commercial products and academic publications advocate the use of AI in the diagnosis and tracking of mental health conditions.[5] However, AI applications based on machine learning operate purely from statistical correlation and lack the situated awareness and ability for nuanced consideration that is required for a responsible diagnosis. Understanding that such diagnostic best practices have been cultivated for good reason, calls attention to the implications of AI applications flouting these contextual norms. For instance, a tendency to privilege "objective" computational diagnoses might lead practitioners to align their clinical judgments with AI systems. This could result in over- or underdiagnosis, as well as negatively affect patient treatment and agency (McCradden et al., 2023). Consequences are likely even more severe when AI tools are marketed directly to the public. Even when merely providing indications rather than a full-blown diagnosis, AI systems can set off or contribute to a spiral of unhelpful self-diagnosis that is often difficult to unmake. Such AI-facilitated self-diagnosis through designated apps or general-purpose chatbots such as ChatGPT threatens the very functioning of mental health care. Patients already assured of their seeming diagnosis and reinforcing the behaviors associated with it, visit mental health professionals often not for a professional diagnosis (itself rendered difficult at this point) but increasingly only for confirmation and medication. The pressing challenges this poses to mental health practitioners illustrate how AI applications can disregard entrenched contextual norms that are central to the very integrity of the field (c.f., Davis, 2022).[6]

**II.2 International Development**

Within international development, moral considerations are discussed in development ethics. And while new macroeconomic strategies (Lipton, 1992), new understandings of poverty (Sen, 1979), and a greater appreciation of colonial legacies (e.g., Langan, 2018) continue to shape the normative landscape of international development, we can also find a considerable body of shared values and norms (Drydyk, 2011) at times codified in codes of ethics (IEDC, 2005).

A central practice in the context of international development is poverty measurement. Over the past century, poverty measurement practice has developed an increasingly rich set of epistemic, methodological, and normative standards. Development practitioners recognize that choices in measurement design and use are inherently normative (Alkire et al., 2015, Chapter 6). Thus, experts engaged in poverty measurement stress the importance of explicitly recognizing motivations that underlie the design and use of poverty metrics and flagging the limitations of their metrics (Ravallion, 2016, pp. 188–190). Relatedly, best practices regarding stakeholder engagement, principled measurement design, and the validation of poverty measurements (Alkire et al., 2015, Chapter 8; Ravallion, 2016, Chapter 5.5) are intended to ensure that poverty measurements hold up to a certain epistemic standard. While such norms are not always lived up to entirely, even as aspirations they guide and coordinate development research and policy.

[5] Among the most speculative applications are those proposing the screening and tracking of conditions such as depression based on social media posts (Eichstaedt et al., 2018), facial expression (Nepal et al., 2024), tone of voice (Larsen et al., 2024), or movement patterns (Müller et al., 2021).

[6] This claim does not deny that part of the current challenges faced in the mental health context (especially regarding resources constraints) also arise due to emerging AI applications lowering barriers to access in ways that should be welcomed.

*AI poverty predictions and methodological norms*

Under the banner of "data for development" machine models are trained on satellite and mobile network data to estimate poverty metrics at low cost and high levels of granularity.[7] In contrast to traditional poverty measurement practice, however, machine learning practitioners commonly fail to validate the metrics their models are trained upon or test the robustness of their estimations (Mussgnug, 2022). Moreover, developers often neglect to emphasize the distinction between the concept of poverty and its operationalization in any given measurement. Doing so collapses the space wherein critical conceptual and normative deliberation take place. In contrast to researchers involved in traditional poverty measurements, machine learning practitioners often do not justify design choices of the measurements they predict nor explicitly lay out how and with which limitations any given metric tries to capture a particular notion of poverty (Mussgnug, forthcoming).

As machine learning applications disregard the methodological norms of poverty measurement, their predictions can fall short of established epistemic standards, providing estimations of questionable evidential standing. The concern is not only speculative. Empirical research demonstrates how AI predictions can fail to track short-term changes in livelihood outcomes (Kondmann & Zhu, 2020), are trained on questionable proxies (Brown et al., 2016), and underestimate cases of extreme poverty (Ratledge et al., 2022). The recent proliferation of cheap AI poverty predictions thus risks compromising the (already imperfect) integrity and reliability of development research. As a result, development resources might be distributed unfairly with potentially disastrous consequences for the communities affected.

*AI poverty predictions and norms of interpretability*

AI predictions might threaten the integrity of international development in a second way. As outlined, poverty measurement practice strives towards rendering transparent design choices in the development of poverty indices. Only when design choices are made explicit and measurements are interpretable can these be considered critically and adopted responsibly for a given purpose. Interpretability is not only important for development practitioners but also for affected communities whose acceptance of interventions and regulations might depend critically on understanding the poverty measurements underlying aid distribution and policymaking. Deep learning models employed for the prediction of poverty metrics, however, can be opaque (Boge, 2022) and, thus, lack interpretability and explainability (Hall et al., 2022). When aid is distributed based on opaque machine learning predictions, affected communities are no longer able to comprehend aid decisions.[8] Consequently, they might lose their ability to meaningfully challenge distributive choices. If expectations regarding transparency and explainability are not met, the use of opaque machine learning models can erode already fragile trust and agency in the international development context.

[7] For an overview, see for instance (Stubbers & Holvoet, 2020).

[8] Even when exaplainable AI methods are employed, their outputs might not be immediately intelligible to affected communities.

### II.3 AI Practice and Contextual Integrity

These examples served to highlight how AI systems can breach not only informational norms but also context-specific epistemic standards, best practices, established procedures, and other contextual norms. As outlined before, violations of entrenched norms are not necessarily morally or epistemically objectionable as in the cases discussed here. However, respect for the collective and cumulative wisdom and practical experience that these norms can reflect calls for a measured conservatism. Contextual norms have often been cultivated over time to promote central epistemic and ethical aims, mitigate potential harm and coordinate actors and their expectations. The cases presented illustrate how the violation of contextual norms and their respective expectations by AI systems can threaten the integrity of social contexts and, as a result, erode trust, inflict harm, and spark justified social outcry. Arguably, the implications are far-reaching also because AI is often employed in contexts such as international development and mental health care whose moral and epistemic integrity is already very fragile.

The point is not to endorse the uncritical adoption of entrenched contextual norms but to underscore the decisive implications of AI's lacking engagement with or even a blanket disregard for established contextual norms. Deviations from such entrenched norms can but also *must* be justified rather than being the result of AI applications being developed and deployed in a manner detached from the social context of application.

## III AI Ethics and the notion of AI as uncharted moral territory

Much research goes into the ethical design and use of AI systems — sometimes in ways that closely engage with the concrete domains of AI applications. At the same time, certain prominent efforts to promote ethical AI can inadvertently contribute to and seemingly legitimize disregard for established contextual norms by endorsing an understanding of AI as a novel and distinct realm for ethical deliberation, norm setting, and virtue cultivation.

Approaches to AI ethics as the exploration of uncharted ethical territory echo a long-standing current in the philosophy of technology. In the 1980s and 1990s, scholars such as James H. Moor (1985) explicitly defended computer ethics as a distinct field of ethical inquiry. Philosophical debate regarding the foundations of computer ethics, however, soon moved on from examining the legitimacy of computer-related issues as a distinct domain of applied ethics. Already assuming the existence of computer ethics as a distinct sphere of moral inquiry, scholars focused on whether the challenges posed by computers necessitate their own radically unique moral philosophy or can be delt with through existing ethical frameworks (for an overview of the debate see Floridi & Sanders, 2002; Tavani, 2010). A similar debate emerged as attention shifted from computer ethics to the internet or "cyberspace" as the next ethical frontier (Tavani, 2005).

The notion of computer and internet ethics as the exploration of uncharted moral territory did not go unchallenged. Donald Gotterbarn, for instance, questioned the tendency to liberally subsume normative considerations under the concept of computer ethics. He remarks rather strikingly: "If these are tales about computer ethics simply because they involve the use of a computer, then my use of a scalpel to rob someone is a problem of medical ethics" (1991, p. 26). Commenting on the regulation of the internet, Helen Nissenbaum calls us to resist "the suggestion that, with regard to privacy, the Net is virgin territory where it falls to the parties to construct terms of engagement for each transaction" (2011, p. 45). Such criticism, however, did little to turn the tide on the notion of emerging technologies as distinct and novel moral territory. Instead, such an understanding of emerging technologies became a persistent undercurrent in the philosophy of technology — spanning from the ethics of computers and the internet to certain strands of AI ethics today.

### III.1 AI Principlism

Consider, for instance, the case of AI principlism. Principlism denotes an approach to AI ethics that prioritizes the formulation and enactment of high-level principles for responsible AI development and use. Principlism emerged as a prominent early framework in AI ethics due to the proliferation of guidelines wherein governments, private corporations, and the third sector self-commit to high-level AI principles. A recent review article surveys no less than two hundred such documents (Corrêa et al., 2023) which converge chiefly around themes of transparency, reliability, justice, privacy, and accountability (ibid; see also Fjeld et al., 2020; Hagendorff, 2020; Jobin et al., 2019).

Principlism takes inspiration from other fields of applied ethics, and in particular biomedical ethics, where scholarly and practical moral deliberation often begins from high-level principles of clinical care (e.g., Beauchamp & Childress, 2013). Underlying AI principlism, thus, is the very idea that an approach to ethics based on principles can be translated from the context of clinical health care to the "context of AI." In this way AI principlism is grounded in and can promote the notion that "Artificial Intelligence" — its development and deployment — somehow constitutes its own moral territory that calls for academics, regulators, and practitioners to cultivate, from the ground up, new designated moral principles.

Such an understanding of AI might not only be attributed to a persistent undercurrent in philosophical scholarship on technology but might also, in part, be motivated by the history of AI itself. Early research on AI centered around the common goal of developing a human or super-human level general intelligence (McCarthy et al., 1995/2006; Wooldridge, 2021). It can be argued that such an autonomous artificial general intelligence, in fact, requires a fundamental rethinking of our normative landscape. Despite the commercial hype around generative AI, the vast majority of work in AI today focuses on the design and deployment of distinct task-specific systems in a wide range of radically heterogeneous contexts (i.e., narrow AI).[9] Nonetheless, narratives of autonomous or even conscious artificial general intelligence popular around the height of principlism (e.g., Bostrom, 2014; Tegmark, 2017) might have played into the notion of AI as new moral territory.

The publication of principle-based guidelines on AI peaked around the year 2018, followed by a sharp decline in scholarly and regulatory interest in AI principles (Corrêa et al., 2023). This is not only due to saturation. With no shortage of AI scandals following the well-marketed publication of AI guidelines, ethicists, regulators, and practitioners have increasingly become disenchanted about the impact and usefulness of high-level ethical principles (Mittelstadt, 2019; Munn, 2023). Today, many AI ethicists agree that a focus on high-level principles has at best been insufficient. At worst, virtue-signaling by self-committing to vague and unenforceable principles has stalled the development of more robust ethical and regulatory approaches (Calo, 2018). Nonetheless, principlism remains a somewhat active strand of research and its legacy (including its framing of AI as new ethical ground) lingers in alternative approaches to AI ethics.

[9] This paper is concerned with precisely those applications at the center of most ethical scholarship and governance efforts today. I do not engage with work on autonomous artificial general intelligence or philosophical research on it.

### III.2 AI Fairness and translational AI Ethics

A particularly influential line of criticism focuses on the actionability of AI principles. Critics note a disconnect between what AI guidelines offer and what developers would find useful — namely a way to translate principles into practice. In the words of Morley et al. (2023, p. 411), the failure of principlism can be attributed in part to the fact that in the form of overarching principles "AI ethics theory remains highly abstract, and of limited practical applicability to those actually responsible for designing algorithms and AI systems." Thilo Hagendorff finds evidence for a lack of engagement with AI practice not only in the absence of on-the-ground technical tools and methods for implementing principles but also in the lack of technical specifics in the language of these guidelines (Hagendorff, 2020, p. 111).

The proposed solution to this issue lies in engaging AI principles with technical practice. Ethicists and AI practitioners must collaborate in developing a more fine grained taxonomy (Morley et al., 2020), technical explanations (Hagendorff, 2020, p. 111), and a comprehensive set of standardized and readily-deployable tools that can help translate principles into technical practice (Morley et al., 2023).[10] In answering the call for more technical specificity, researchers have put a particular emphasis on engineering technical methods for designing fairer AI models. Early research on AI fairness promised to develop universal mathematical operationalizations of fairness and off-the-shelf statistical methods to mitigate discrimination against relevant subgroups. This approach to AI fairness, however, has come under criticism for its understanding of fairness as a technical property of an AI model itself rather than a property of the social (or sociotechnical) systems within which AI is employed. By focusing only on AI as a technical system, early research on AI fairness failed to acknowledge how different notions of fairness are operative in different social contexts and how fairness concerns not machine learning predictions themselves but the outcomes of decision processes involving both social actors and AI (Dolata et al., 2022; Selbst et al., 2019).[11]

Again, we can see in this early technology-centered approach to AI fairness the tendency to understand AI as its own distinct moral territory. Ethical considerations regarding fairness were seen as issues pertaining to the "domain of AI" rather than use of AI in diverse social contexts. Researchers aimed to statistically conquer ethical and social challenges related to the use of AI while failing to engage with the contextual norms around fairness that govern the radically heterogenous social domains within which AI is employed.

AI fairness is also the central concern of an influential article by Abigail Z. Jacobs and Hanna Wallach (2021) on machine learning and measurement modeling. Jacobs and Wallach argue that AI practitioners often conflate concepts with their operationalization found in data.[12] This, however, renders difficult the identification of potential mismatches between a concept (e.g., depression) and its operationalization (e.g., a particular psychometric test of depression) at the root of many fairness-related harms.[13] Measurement modeling, common in the quantitative social sciences, can help identify these mismatches . Thus, the authors advocate for translating methodological norms and best practices such as measurement modeling and construct validation from social measurement to AI. Their interdisciplinary approach not only calls attention to a central shortcoming of AI practice but also directs practitioners to relevant methodological resources.

[10] One might wonder to what extent such is helpful in relation to the problems presented in section two if technical AI practice is itself the locus of problematic abstraction.

[11] A similar criticism might be leveraged against certain parallel approaches to explainable AI.

[12] Jacobs and Wallach propose to use measurement modeling and construct validation both as a methodological resource or AI practitioners and to better understand debates within AI fairness itself. My criticism extends only to the former and focuses on the framing of the argument not the substance (which I am very sympathetic to).

[13] Jacobs and Wallach use the term "constructs" popular in social science research and especially in psychometrics.

At the same time, the very framing of *translating* norms such as measurement modeling and construct validation from the realm of social measurement to AI can be seen as yet another example of approaching AI as a distinct normative realm. AI ethicists translating entrenched norms from contexts of practice within which AI is employed to the "realm of AI" might be accused of moral Columbusing — discovering anew norms that have long been cultivated and refined in domains of AI application rather than understanding AI systems as new elements within the practice of social quantification already accountable to its methodological norms (Mussgnug, 2022).

**III.3 AI Ethics and Moral Philosophy**

The notion of AI as its own distinct ethical territory can also be encountered in certain scholarship applying general frameworks from moral and political philosophy to AI ethics. For instance, some moral philosophers (often aligned with the effective altruism movement) have embraced versions of utilitarianism as a means to map out the novel moral territory of AI (Srinivasan, 2015). The notion of AI as a distinct moral domain is illustrated, for example, in an article by Štěpán Cvik which disentangles approaches of utilitarianism and their respective challenges "*in the context of artificial intelligence*" (2022, p. 291, emphasis added).

Others have criticized the utilitarian approach to AI ethics and relied instead, for instance, on Rawls' theory of justice (for an overview see Bay, 2023). Hereby it is important to distinguish those that leverage Rawlsian ethics to explore and assess the role of AI in relation to public institutions or the basic structure of society (e.g., Gabriel, 2022; Grace & Bamford, 2020) from those that employ Rawls' theory of justice to the "realm of AI." As an example of the latter, Salla Westerstrand (2024) proposes to ground ethical AI guidelines in Rawls' theory of justice. Applications of Rawlsian ethics to AI in the context of public institutions might be interpreted as an attempt to integrate AI within existing or aspirational normative structures in the realm of public institutions. Those attempting to pioneer Rawls' theory of justice in the general domain of AI, however, provide no less evidence for the persistent interpretation of AI as a distinctive and novel moral territory than some scholars advocating for AI utilitarianism.

**III.4 AI as Uncharted Ethical Territory and Contextual Integrity**

AI principlism, technical approaches to AI fairness, certain translational research, or some attempts to mobilize general moral philosophies for 'the context of AI' are not the only instances in AI ethics that illustrate a tendency to understand AI as a virgin moral domain. A similar inclination can be observed in approaches to virtue ethics that emphasize the cultivation of new technomoral virtues (c.f., Vallor, 2016) without engaging existing contextual standards of excellence, or can be encountered in pursuits of generic AI safety certifications that fail to reference entrenched contextual norms cultivated to promote reliable practice in particular areas (Corrêa & Mönig, 2024; Winter et al., 2021).

Granted, some approaches in AI ethics are not concerned with the moral implications of AI more broadly but focus explicitly and narrowly on the professional context of AI development. For instance, a small fraction of guidelines are framed specifically as ethical principles of AI development (e.g., Blackman, 2020) and some scholars explore the role of professional norms in AI engineering (e.g., Gasser & Schmitt, 2020) or which ethical and epistemic virtues should guide AI practitioners (e.g., Hagendorff, 2022). Embracing the professionalization of AI development and instilling in practitioners a professional identity centered first and foremost around the shared technical dimensions of their work might, in its own ways, contribute to disengagement with the diverse contexts of AI application. Addressing AI ethics primarily as an occupational ethics also faces other problems. For instance, fostering responsible AI requires not only a professional ethics for individual developers but also a broader organizational ethics (Mittelstadt, 2019). Moreover, the emergence of no-code AI platforms renders AI development accessible even to laymen and ethical concerns often center around the widespread use of AI by non-experts, which are outside the focus of occupational AI ethics. For these and other reasons, approaches to AI ethics exclusively as an occupational ethics only make up a small share of the scholarship.[14]

In any case, this section does not issue a blanket critique of the current state of AI ethics. Instead, it aims to illustrate how a tendency to approach emerging technologies as uncharted moral territories extends from early research on computer and internet ethics to certain prominent strands of AI ethics today. My central claim is that this notion of AI as uncharted ethical territory can play into AI practitioners' disregard for entrenched contextual norms and, as a result, contribute to AI applications threatening contextual integrity. It does so in at least two ways. First, by shifting the focus of ethical debate narrowly toward moral innovation and, second, by seemingly legitimizing disregard for existing contextual normative structures.

The notion of AI as its own distinct uncharted normative ground can lead AI ethics to exclusively or disproportionately focus on moral innovation. Much emphasis is being placed on creating and promoting new principles, new virtues, new epistemic standards, new methodological best practices, and new norms. This approach, however, can come at the expense of ethicists engaging with the established norms that already govern the many diverse contexts of AI application. An understanding of AI as uncharted moral territory can, thus, lead scholars to deliberate about the ethical implications and governance of AI without first thoroughly taking stock of the distinctive normative structures already in place in various domains of AI application. In these cases, AI ethicists act little different from some ecologists and development experts upheaving or "innovating" existing agricultural systems in the Global South without understanding and respecting how indigenous practices of cultivation have often gradually developed in ways attuned to the particularities of their land and society (c.f., Belay & Mugambe, 2021). And much like these often disastrous approaches to agricultural reform threaten the integrity of local communities, a narrow focus on moral innovation can threaten the integrity of diverse contexts of AI application.

Arguably, this focus on normative innovation is also reflected in certain legislative approaches such as the recent EU AI Act (2024) that seek to establish novel cross-sectoral regulatory landscapes. Attempts to innovate distinct and overarching regulations for general purpose technologies such as the internet should serve as a cautionary tale regarding the viability of this approach to the governance of AI. In fact, Helen Nissenbaum motivates her framework of contextual integrity, in part, by linking catastrophic privacy scandals to treating general-purpose technologies such as the internet as virgin territory and as a general context for regulation. She notes that her approach "does not support substantive prescriptions for general families of technologies, […], although they ought to be carefully analyzed in terms of the types of powers they offer for affecting (sometimes in extreme ways) the flow of information. Rather, the most fruitful assessments take place within particular contexts […]." (Nissenbaum, 2009, p. 200).

Most central to the purpose of this paper, however, is how the focus on normative innovation in AI ethics and governance can play into the tendency of AI practitioners to disregard existing contextual norms. As public, regulatory, and academic debate centers around the promotion of new principles, virtues, best practices and norms for the uncharted ethical territory of AI, even those developers and users engaging closely with debates around the ethics and governance of AI can all too easily lose sight of existing normative structures in the many areas of AI application.

[14] See also a parallel debate in computer ethics (Gotterbarn, 1991).

The notion of AI as a distinct moral territory also more directly legitimizes the tendency of AI practitioners to abstract from established contextual norms. If AI applications were to be firmly understood as novel elements within existing social contexts and their respective normative structures, developers would already be accountable to its best practices, established procedures, methodological and epistemic requirements, standards of excellence, cultivated virtues, and other contextual norms. Conversely, if AI is framed as its distinct ethical realm, these contextual norms become external to AI practice. In this case, they require separate justification and AI practitioners might easily interpret adherence to them as an optional effort rather than an inherent obligation. AI practitioners might be excused from ignoring norms not yet translated to AI — even if this can threaten the very integrity of social context.

## IV Integrative AI Ethics

This raises the question to what extent AI ethics requires a reorientation — a shift in emphasis from prioritizing the advancement of new AI-centered principles, norms, and virtues to the responsible and considered integration of AI within established social contexts and their respective normative structures. Such an *integrative AI ethics* emphatically identifies AI applications as first and foremost elements within existing contexts of practice — as AI applications *within* rather than *to* mental health care, international development, agriculture, or education.

An integrative approach to AI ethics does not dispute that technology-specific considerations are central to AI ethics, nor does it entirely deny the need for cross-contextual moral theorizing and ethical governance. But it calls for a shift in emphasis grounded in the acknowledgement that, as many have noted, the moral implications of AI often play out differently in the radically heterogenous contexts within which AI finds application. Adopting the words of Thilo Hagendorff (2020, pp. 114–115), the goal is to cultivate an AI ethics that deals less with AI as such.

My call for an integrative AI ethics is aligned and seeks to capture a broader trend toward greater contextual engagement in AI practice, regulation, and moral debate. For instance, increasingly many AI practitioners have formed close and lasting interdisciplinary collaborations with domain experts. Moreover, many degree programs now equip domain experts with technical AI skills resulting in a greater share of professionals with both extensive domain knowledge and AI expertise.[15] Both developments are helping to bring context-specific considerations to the forefront in AI development. Context-specific concerns have also received greater attention in recent regulatory approaches. The UK's 2024 framework for regulating AI, for instance, stresses its orientation toward the use of AI rather than the technology itself (UK Government, 2024). As the authors note: "Rather than target specific technologies, it [the regulatory framework] focuses on the contexts in which AI is deployed" (UK Government, 2023).

Within AI ethics, ethnographic research (often from scholars in science and technology studies) engages in detail with how novel AI systems operate in distinct social contexts. Researchers have explored, for instance, the ethics of AI in migration management (e.g., Molnar, 2021) or higher education (Noteboom, forthcoming.; Noteboom & Ross, 2024). Relatedly, a recent special issue in *AI & Society* advances work on "embedding AI in society" (e.g., Pflanzer et al., 2023). Particularly noteworthy is the growing scholarship on sociotechnical AI ethics. As outlined earlier, sociotechnical research problematizes how practitioners often conceive of AI systems in narrowly technical terms. In response, scholars stress the need to consider the complex entanglement of social and technological aspects in AI. Sociotechnical scholarship also explores the close relationship between technology and morality. Under the heading of technomoral change, researchers emphasize how the introduction of new technologies can alter existing moral beliefs and practices (Johnson & Verdicchio, 2024; Nickel et al., 2022; Swierstra et al., 2009).

[15] Take for instance degree programs in AI and bioscience (e.g., *Artificial Intelligence in the Biosciences MSc - Queen Mary University of London*, n.d.) or economics (Robinson, 2023).

Integrative AI ethics, following Nissenbaum's framework of contextual integrity, goes beyond merely attending more closely to social contexts of application, by advocating for a moderately conservative approach to such efforts. It proceeds from a presumption in favor of entrenched norms based on the assumption that those are often the result of cultivated wisdom, accumulated experience, and a settled rationale within a social context. AI ethicists are asked to put into focus and hold AI practitioners accountable to context-specific best practices, established procedures, methodologies, standards of excellence, and other contextual norms. Rather than highlighting the disruptive potential of AI, ethicists should stress the need to honor expectations grounded in entrenched contextual norms absent convincing reasons to the contrary. These can involve, for instance, psychiatric patients' expectations of confidentiality, mental health practitioners' expectations regarding diagnostic standards, development economists' expectations concerning methodological best practices in poverty measurement or affected communities' expectations in relation to the interpretability of aid decisions.

**IV.1 Objections**

My call to refocus AI ethics along contextual norms may provoke concerns regarding its feasibility and implications. In what follows, I seek to briefly address four possible objections.

First, one might be concerned that this approach results in a problematic degree of conservatism in the moral debate around AI. I believe that such a concern is warranted. Thus, it is important to stress that integrative AI ethics calls for a *measured and conditional* conservatism. It calls for a prima facie respect for entrenched contextual norms, not deference. It does not deny that existing norms can be misguided and that the introduction of AI tools might challenge existing rationales. As mentioned, we might sometimes employ new technologies precisely with the intention to overhaul existing customs. In many areas, the introduction of AI technologies provides us with the timely opportunity to reconsider objectionable conventions and norms often based on racism, sexism, classism, and colonial legacies. If developed prudently, AI can help us pursue more equitable futures and greater human flourishing. Integrative AI ethics emphasizes that for those efforts to be most impactful, they must first be justified with reference to entrenched contextual norms and their shortcomings, as well as promoted in ways coordinated with a context's wider normative structure.

Second, and relatedly, one might worry that an integrative AI ethics is bound to be too descriptive, lacking a certain degree of normativity central to AI ethics — a concern that has been raised with respect to STS scholarship (Mason-Wilkes, 2024) or the empirical turn in the philosophy of technology (Scharff, 2012). I believe such an objection is somewhat misguided. An integrative approach to AI ethics might indeed take strong normative stances on the ethical and socially relevant dimensions of AI. It differs only with respect to the source of such normativity. It argues that normative direction should come *to a lesser extent* from philosophers' moral innovation and, instead, underscores the cultivated wisdom and accumulated moral and practical experience already embodied in existing contextual norms as a significant source of normativity.

Third, one might wonder whether the recent popularization of foundation models such as large language models (LLMs) poses challenges to a context-centered approach to the ethics and governance of AI.

Foundation models are trained on a large corpus of unlabeled data and can be used (often with minimal fine-tuning) for a wide range of purposes across diverse contexts (European Union, 2023). Thus, the question emerges which contextual norms should guide the development and deployment of foundation models?

Rather than pose challenges to integrative AI ethics, I believe that the latter's focus on contextual norms can help bring out clearly the broad responsibilities that are associated with the development and deployment of foundation models. The fact that foundation models are commonly employed across contexts does not absolve practitioners from contextual norms. Instead, it highlights that practitioners must engage with the respective norms that govern any given context in which foundation models find application. If foundation models are used in education, health care, finance, or the legal context, then systems must respect (or practitioners must substantially justify their deviation from) that context's normative structures. Such is clearly a tall order but not unreasonable.

Let us, *only for the purpose of this argument*, buy into the anthropomorphizing narrative of foundation models such as LLMs approaching human or superhuman-level artificial general intelligence (c.f., Y Combinator, 2024). Human actors often engage in a wide range of social contexts. One might work in the education, financial, medical, or legal sector, meet friends at their home, go to church on Sundays, engage in politics, and so on. This does not absolve us from the distinct norms that govern these contexts but renders us accountable to them. It requires us to navigate them thoughtfully, adapting our behavior to align with the specific expectations, responsibilities, and ethical standards each context entails. We should ask no less of foundation models seemingly approaching human-level capabilities.

How exactly to engineer and deploy foundation models to fulfill this task is undoubtedly a difficult question. Practitioners might design and implement models with mechanisms that allow them to adapt to diverse contexts, fine-tune distinct context-specific models, or prohibit certain uses of foundation models. Recent efforts highlight that we likely need a combination of these and other strategies (Deng et al., 2024; Longpre et al., 2024). Integrative AI ethics underscores that central to this emerging moral debate should not be the innovation of norms for the new "realm of foundation models" but engagement with the normative structures governing the diverse contexts of their application.

Lastly, one might question whether the explicit call for an "integrative AI ethics" is needed given the already ongoing push toward greater contextual engagement outlined before. In response, it is important to acknowledge that AI ethics is also a terrain of contestation where much current debate centers around what AI ethics entails and who gets to define it (Green, 2021). The purpose of this paper and of the concept of integrative AI ethics is to explicitly position ongoing context-sensitive work in opposition to an influential understanding of AI as uncharted moral territory and the problems stemming from it, while advocating for a moderately conservative approach to such efforts.

This also serves to sensitize us to the ways in which a notion of AI as radically novel moral terrain can re-enter even efforts grounded in closer engagement with the social contexts of AI application. For instance, a small part of sociotechnical AI scholarship enthusiastically buys into the hype of radical AI disruption (Duarte et al., 2024) and focuses narrowly on how AI systems constitute fundamentally new sociotechnical and technomoral systems (e.g., Accoto, 2024; Bisconti et al., 2024). In this way, the notion of AI as novel moral territory and resulting disengagement with existing contextual norms can enter even sociotechnical scholarship on AI.

The fact that sociotechnical scholarship, founded upon the importance of social factors for AI, is not immune to the notion of AI as uncharted moral territory cannot be attributed to the prevalence of this view within technology ethics alone. Instead, we must also consider how such an understanding is strikingly opportune to both AI practitioners and AI ethicists. The notion of AI as a discrete moral territory conveniently frees practitioners from engaging with the often-extensive norms governing areas of AI application. As Selbst et al. (2019, p. 60) rightly note the tendency to abstract from the contexts of application facilitates the rapid proliferation of machine learning in the first place. It is this abstraction which grounds the idea that a developer can one day develop medical AI, the next day consult on an AI system in accounting, and simultaneously help developing AI "solutions" for social good. But the understanding of AI as virgin ethical territory is no less opportune to the AI ethicist. It calls for the philosopher or ethically minded researcher to "conquer" the ethical terra nullius of AI — a prospect that no doubt flatters our disciplinary ego. The ethicist, not beholden to distinctive moral expertise in any one of the radically heterogenous contexts in which AI is employed, can simultaneously consult on the responsible development and use of AI across health care, scientific research, finance, education, and so on. In the best case, engagement with entrenched contextual norms then takes on the form of philosophical Columbusing, boosting our publication statistics by translating established contextual norms to the realm of AI. In the worst case, AI ethics lacks engagement with contextual best practices, procedures, methodologies, virtues, standards of excellence, and other norms entirely. As argued, AI ethics can then contribute to practitioners' disengagement with contextual norms. And this threatens the integrity of social contexts in ways that are often central to why AI systems continue to defy expectations, erode trust, and spark societal outcry.

## V Conclusion

Recent research illustrates how AI development and deployment can happen in a manner detached from the concrete social context of application. This paper has emphasized how, by abstracting from the contexts of AI application, practitioners also disengage from the distinct normative structures that govern them. This disregard for contextual norms can threaten the integrity of a context with often decisive ethical implications. Certain prominent efforts to promote responsible and ethical AI can inadvertently contribute to and seemingly legitimize this disregard for established contextual norms. Echoing a persistent undercurrent in technology ethics of understanding emerging technologies as uncharted moral territory, current approaches to AI ethics can promote a notion of AI as a novel and distinct realm for ethical deliberation, norm setting, and virtue cultivation. A narrative of AI as virgin moral territory, however, can come at the expense of practitioners, policymakers and ethicists engaging with already established norms and virtues that were gradually cultivated to promote successful and responsible practice within diverse social contexts. In response, this paper questioned the current narrow prioritization in AI ethics of moral innovation over moral preservation and advocated for a moderately conservative approach to the ethics of AI that prioritizes the responsible and considered integration of AI within established social contexts and their respective normative structures.

## Acknowledgement

I would like to thank Shannon Vallor for her generous support in writing this article. I am grateful to Fabio Tollon, as well members of NC State's NeuroComputational Ethics Group and the Centre for Technomoral Futures at the University of Edinburgh for their feedback. Work on this paper was supported by the Baillie Gifford Scholarship in AI Ethics at the Centre for Technomoral Futures.